\documentclass[10pt,twocolumn,letterpaper]{article}

\usepackage{cvpr}              % To produce the CAMERA-READY version
\usepackage{graphicx}
\usepackage{amsmath}
\usepackage{amssymb}
\usepackage{booktabs}
\usepackage{xcolor,colortbl}

\usepackage[pagebackref,breaklinks,colorlinks]{hyperref}

\makeatletter
\renewcommand{\paragraph}{%
  \@startsection{paragraph}{4}%
  {\z@}{0ex \@plus 1ex \@minus .2ex}{-1em}%
  {\normalfont\normalsize\bfseries}%
}
\makeatother

\usepackage[capitalize]{ cleveref}
\crefname{section}{Sec.}{Secs.}
\Crefname{section}{Section}{Sections}
\Crefname{table}{Table}{Tables}
\crefname{table}{Tab.}{Tabs.}

\def\cvprPaperID{1200} % *** Enter the CVPR Paper ID here
\def\confName{CVPR}
\def\confYear{2023}

\usepackage{array}
\newcolumntype{P}[1]{>{\centering\arraybackslash}p{#1}}
\newcolumntype{M}[1]{>{\centering\arraybackslash}m{#1}}

\begin{document}
\def\Blue{\color{blue}}
\def\Purple{\color{purple}}

\def\A{{\bf A}}
\def\a{{\bf a}}
\def\B{{\bf B}}
\def\b{{\bf b}}
\def\C{{\bf C}}
\def\c{{\bf c}}
\def\D{{\bf D}}
\def\d{{\bf d}}
\def\E{{\bf E}}
\def\e{{\bf e}}
\def\f{{\bf f}}
\def\F{{\bf F}}
\def\K{{\bf K}}
\def\k{{\bf k}}
\def\L{{\bf L}}
\def\H{{\bf H}}
\def\h{{\bf h}}
\def\G{{\bf G}}
\def\g{{\bf g}}
\def\I{{\bf I}}
\def\R{{\bf R}}
\def\X{{\bf X}}
\def\Y{{\bf Y}}
\def\OO{{\bf O}}
\def\oo{{\bf o}}
\def\P{{\bf P}}
\def\p{{\bf p}}
\def\Q{{\bf Q}}
\def\r{{\bf r}}
\def\s{{\bf s}}
\def\S{{\bf S}}
\def\t{{\bf t}}
\def\T{{\bf T}}
\def\x{{\bf x}}
\def\y{{\bf y}}
\def\z{{\bf z}}
\def\Z{{\bf Z}}
\def\M{{\bf M}}
\def\m{{\bf m}}
\def\n{{\bf n}}
\def\U{{\bf U}}
\def\u{{\bf u}}
\def\V{{\bf V}}
\def\v{{\bf v}}
\def\W{{\bf W}}
\def\w{{\bf w}}
\def\0{{\bf 0}}
\def\1{{\bf 1}}
\def\N{{\bf N}}

\def\AM{{\mathcal A}}
\def\EM{{\mathcal E}}
\def\FM{{\mathcal F}}
\def\TM{{\mathcal T}}
\def\UM{{\mathcal U}}
\def\XM{{\mathcal X}}
\def\YM{{\mathcal Y}}
\def\NM{{\mathcal N}}
\def\OM{{\mathcal O}}
\def\IM{{\mathcal I}}
\def\GM{{\mathcal G}}
\def\PM{{\mathcal P}}
\def\LM{{\mathcal L}}
\def\MM{{\mathcal M}}
\def\DM{{\mathcal D}}
\def\SM{{\mathcal S}}
\def\RB{{\mathbb R}}
\def\EB{{\mathbb E}}

\def\tx{\tilde{\bf x}}
\def\ty{\tilde{\bf y}}
\def\tz{\tilde{\bf z}}
\def\hd{\hat{d}}
\def\HD{\hat{\bf D}}
\def\hx{\hat{\bf x}}
\def\hR{\hat{R}}

\def\Ome{\mbox{\boldmath$\omega$\unboldmath}}
\def\bet{\mbox{\boldmath$\beta$\unboldmath}}
\def\et{\mbox{\boldmath$\eta$\unboldmath}}
\def\ep{\mbox{\boldmath$\epsilon$\unboldmath}}
\def\ph{\mbox{\boldmath$\phi$\unboldmath}}
\def\Pii{\mbox{\boldmath$\Pi$\unboldmath}}
\def\pii{\mbox{\boldmath$\pi$\unboldmath}}
\def\Ph{\mbox{\boldmath$\Phi$\unboldmath}}
\def\Ps{\mbox{\boldmath$\Psi$\unboldmath}}
\def\pss{\mbox{\boldmath$\psi$\unboldmath}}
\def\tha{\mbox{\boldmath$\theta$\unboldmath}}
\def\Tha{\mbox{\boldmath$\Theta$\unboldmath}}
\def\muu{\mbox{\boldmath$\mu$\unboldmath}}
\def\Si{\mbox{\boldmath$\Sigma$\unboldmath}}
\def\Gam{\mbox{\boldmath$\Gamma$\unboldmath}}
\def\gamm{\mbox{\boldmath$\gamma$\unboldmath}}
\def\Lam{\mbox{\boldmath$\Lambda$\unboldmath}}
\def\De{\mbox{\boldmath$\Delta$\unboldmath}}
\def\vps{\mbox{\boldmath$\varepsilon$\unboldmath}}
\def\Up{\mbox{\boldmath$\Upsilon$\unboldmath}}
\def\Lap{\mbox{\boldmath$\LM$\unboldmath}}
\newcommand{\ti}[1]{\tilde{#1}}

\def\tr{\mathrm{tr}}
\def\etr{\mathrm{etr}}
\def\etal{{\em et al.\/}\,}
\newcommand{\indep}{{\;\bot\!\!\!\!\!\!\bot\;}}
\def\argmax{\mathop{\rm argmax}}
\def\argmin{\mathop{\rm argmin}}
\def\vec{\text{vec}}
\def\cov{\text{cov}}
\def\dg{\text{diag}}

\newcommand{\tabref}[1]{Table~\ref{#1}}
\newcommand{\secref}[1]{Sec.~\ref{#1}}
\newcommand{\figref}[1]{Fig.~\ref{#1}}
\newcommand{\lemref}[1]{Lemma~\ref{#1}}
\newcommand{\thmref}[1]{Theorem~\ref{#1}}
\newcommand{\clmref}[1]{Claim~\ref{#1}}
\newcommand{\crlref}[1]{Corollary~\ref{#1}}
\newcommand{\eqnref}[1]{Eqn.~\ref{#1}}

\newtheorem{remark}{Remark}
\newtheorem{theorem}{Theorem}
\newtheorem{lemma}{Lemma}
\newtheorem{definition}{Definition}

\newtheorem{proposition}{Proposition}

\newcommand{\cz}[1]{{\color{blue}[Chengzhi says: #1]}}
\newcommand{\rl}[1]{{\color{red}[Ruoshi says: #1]}}

\title{Humans as Light Bulbs: 3D Human Reconstruction from Thermal Reflection}

\newcommand{\carl}[1]{\textcolor{red}{Carl:#1}}
\newcommand{\ruoshi}[1]{\textcolor{blue}{Ruoshi:#1}}

\definecolor{object}{rgb}{0.96, 0.32, 0.50}
\definecolor{human}{rgb}{0.47, 0.60, 0.92}
\definecolor{Gray}{gray}{0.9}

\author{Ruoshi Liu and Carl Vondrick
\vspace{0.1cm}
\\Columbia University
\vspace{0.06cm}
%{\tt\small \{rliu\}@cs.columbia.edu}\\
\\\href{https://thermal.cs.columbia.edu/}{\textbf{\url{thermal.cs.columbia.edu}}}
}

\twocolumn[{%
        \renewcommand\twocolumn[1][]{#1}%
        \maketitle
        \begin{center}
                \vspace{-0.5cm}
                \includegraphics[width=1.0\linewidth]{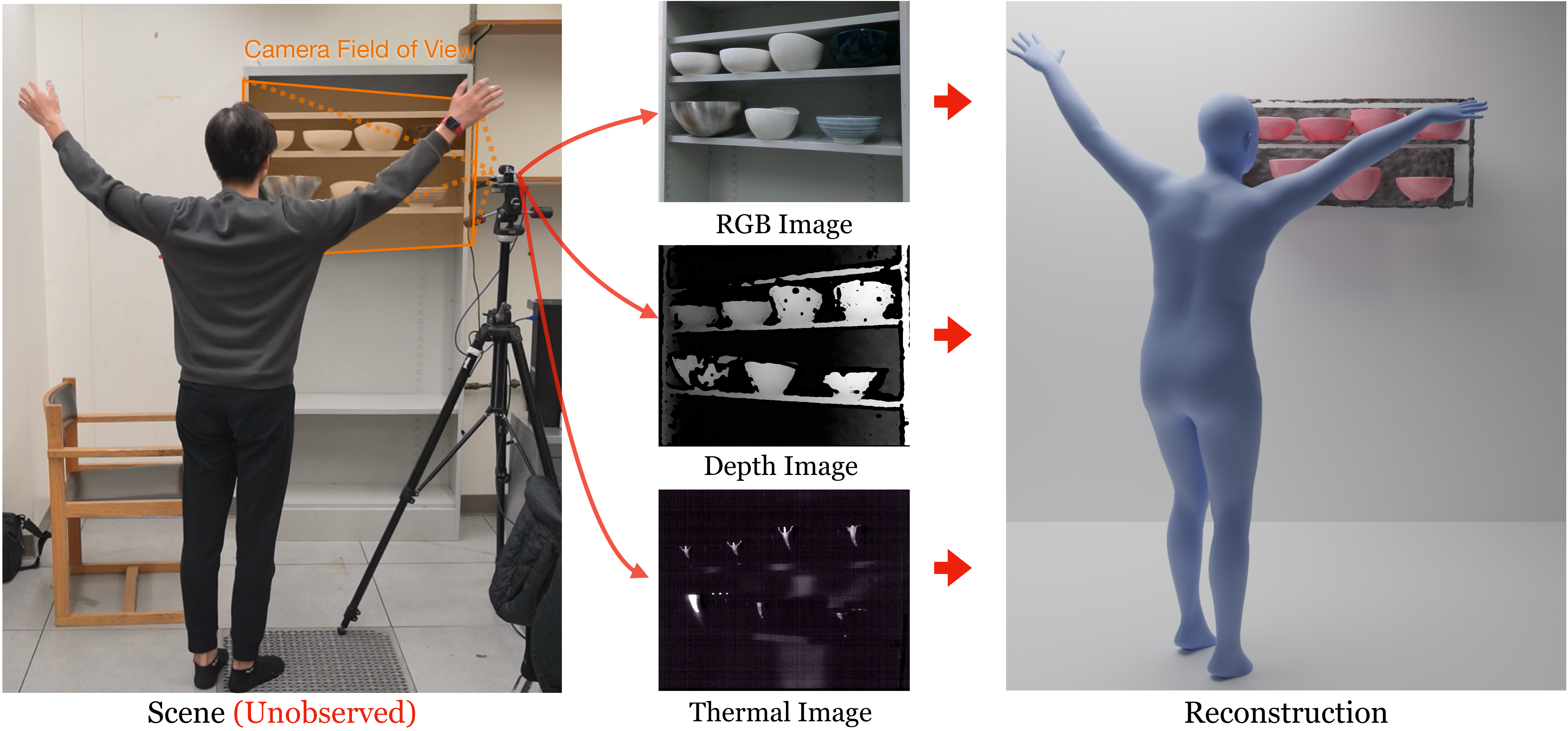}
                \captionof{figure}{We introduce a method to reconstruct the 3D position and pose of a person from their thermal reflections in everyday objects with non-planar surfaces. Given an RGBD image and a thermal image in the middle of the figure, our method is able to recover the 3D mesh of the person (\textcolor{human}{blue}) as well as the objects (\textcolor{object}{pink}), even though they are not within the field of view of the camera system. Our system \textbf{never sees the scene on the left}, which is only shown for visualization purposes.
                \vspace{1em}
                }
                \label{fig:teaser}
        \end{center}
}]

% \begin{figure*}[t]
%     \centering
%     \includegraphics[width=\linewidth]{figures/teaser.pdf}
%     \caption{The setup of our experiments. }
%     \label{fig:tiser}
% \end{figure*}

%%%%%%%%% ABSTRACT
\begin{abstract}
\vspace{-0.5em}
The relatively hot temperature of the human body causes people to turn into long-wave infrared light sources. Since this emitted light has a larger wavelength than visible light, many surfaces in typical scenes act as infrared mirrors with strong specular reflections. We exploit the thermal reflections of a person onto objects in order to locate their position and reconstruct their pose, even if they are not visible to a normal camera. We propose an analysis-by-synthesis framework that jointly models the objects, people, and their thermal reflections, which combines generative models with differentiable rendering of reflections. Quantitative and qualitative experiments show our approach works in highly challenging cases, such as with curved mirrors or when the person is completely unseen by a normal camera.

%Thermal reflection has been a ``bug" in thermal camera, causing inaccurate temperature measurement of specular surfaces. We leverage this ``bug" and the fact that humans are light sources in thermal camera to solve the dual problems of human and object reconstruction in 3D with a single camera view. By integrating generative models and differentiable ray tracing, we perform optimization of humans and objects in 3D to match the observed thermal reflection in the objects. Experiments show that our method can accurately reconstruct the location, orientation, and shape of both human and object by looking at the thermal reflection of human in the object, which generalizes across experiments in simulation and real world.

\end{abstract}

%%%%%%%%% BODY TEXT
\section{Introduction}
One of the major goals of the computer vision community is to locate people and reconstruct their poses in everyday environments. What makes thermal cameras particularly interesting for this task is the fact that humans are often the hottest objects in indoor environments, thus becoming infrared light sources. Humans have a relatively stable body temperature of 37 degrees Celcius, which according to the Stefan-Boltzmann law, turns people into a light source with constant brightness under long-wave infrared (LWIR). This makes LWIR images a robust source of signals of human activities under many different light and camera conditions.

Since infrared light on the LWIR spectrum has a wavelength that is much longer than visible light (8$\mu$m-14$\mu$m vs.\ $0.38\mu$m-$0.7\mu$m), the objects in typical scenes look qualitatively very different from human vision. Many surfaces of objects in our daily life -- such as a ceramic bowl, a stainless steel fridge, or a polished wooden table top -- have stronger specular reflections than in the visible light spectrum\cite{oren1994generalization, bennett1961relation}. Figure \ref{fig:teaser} shows the reflection of a person with the surface of salad bowls, which is barely visible to the naked eye, if at all, but clearly salient in the LWIR spectrum.

In cluttered environments, a visible light camera may not always be able to capture the person, such as due to a limited field of view or occlusions. 
In such scenes, the ideal scene for locating and reconstructing a person would be an environment full of mirrors.  This is what the world looks like under the LWIR spectrum.
Infrared mirrors are abundant in the thermal modality, and 
reflections reveal significant non-line-of-sight information about the surrounding world.  

In this paper, we introduce a method that uses the image of a thermal reflection in order to reconstruct the position and pose of a person in a scene. We develop an analysis-by-synthesis framework to model objects, people, and their thermal reflections in order to reconstruct people and objects. Our approach combines generative models with differentiable rendering to infer the possible 3D scenes that are compatible with the observations. Given a thermal image, our approach optimizes for the latent variables of generative models such that light emitting from the person will reflect off the object and arrive at the thermal camera plane.

Our approach works in highly challenging cases where the object acts as a curved mirror. Even when a person is completely unseen by a normal visible light camera, our approach is able to localize and reconstruct their 3D pose from just their thermal reflection. 
Traditionally, the increased specularity of surfaces has posed a challenge to thermography, making it extremely difficult to measure the surface temperature of a thermally specular surface, which brings out a line of active research aiming to remove the specular reflection for more accurate surface temperature measurement \cite{batchuluun2020region, li2018removal, batchuluun2019study, zeise2016temperature}. We instead exploit these ``difficulties'' of LWIR to tackle the problem of 3D human reconstruction from a single view of thermal reflection image.

The primary contribution of the paper is a method to use the thermal reflection of the human body on everyday objects to infer their location in a scene and its 3D structure. The rest of the paper will analyze this approach in detail. Section \ref{related} provides a brief overview of related work for 3D reconstruction and differentiable rendering. Section \ref{method} formulates an integrated generative model of humans and objects in a scene, then discusses how to perform differentiable rendering of reflection, which we are able to invert to reconstruct the 3D scene. Section \ref{experiments} analyzes the capabilities of this approach in the real world. We believe thermal cameras are powerful tools to study human activities in daily environments, extending computer vision systems' ability to function more robustly even under extreme light conditions.
\looseness-1

\begin{figure*}[t]
    \centering
    \includegraphics[width=\linewidth]{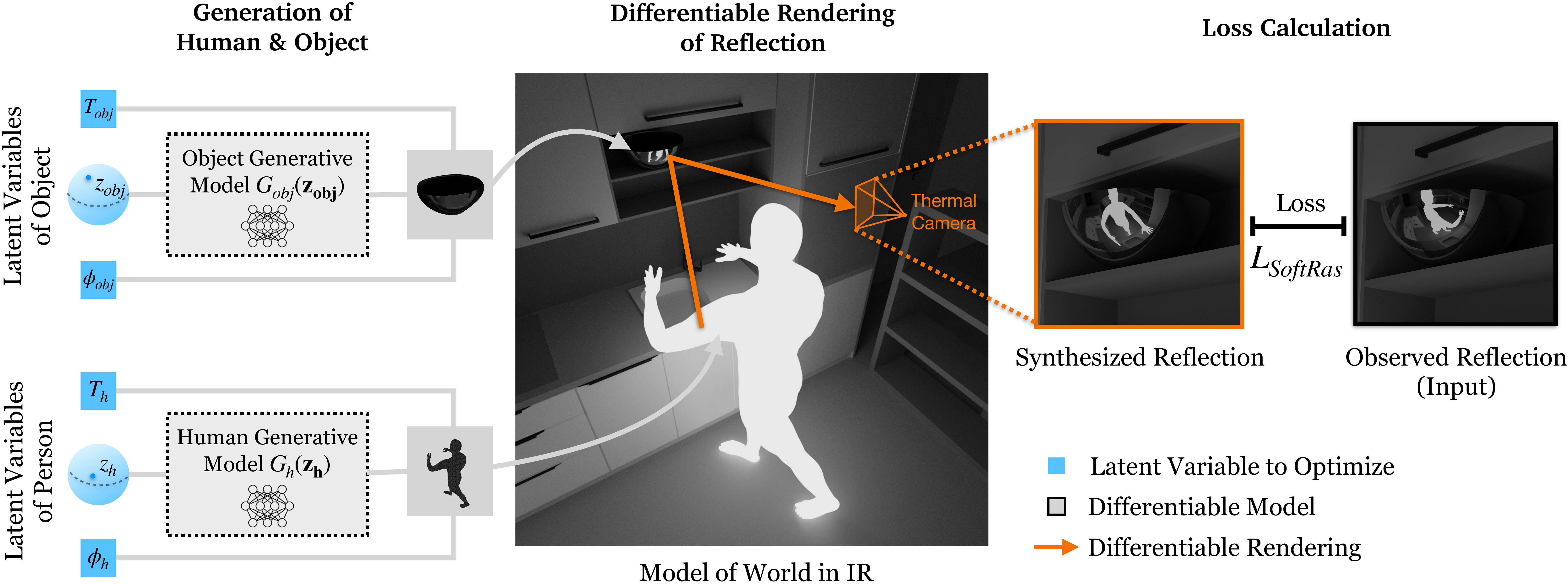}
    \caption{High-level overview of our analysis-by-synthesis framework. We sample random initializations from the latent space of pretrained generative models of humans and objects in 3D. Through a differentiable rendering process, we synthesize a reflection image of a human body on object surfaces. This synthesized reflection is compared with the observed reflection with an $L_1$ loss. Gradients are backpropagated through differentiable rendering and generative models to the latent variables.}
    \vspace{-0.2cm}
    \label{fig:method}
\end{figure*}

\section{Related Work} \label{related}

\textbf{Differentiable Rendering.}
Differentiable rendering is a differentiable process of rendering 2D images given 3D scenes. The gradient obtained from the image space w.r.t.\ the scene parameters can be calculated and used to perform optimization. Recent advances in implicit 3D representations, especially Neural Radiance Field (NeRF) \cite{mildenhall2021nerf, barron2021mip, barron2022mip, muller2022instant, srinivasan2021nerv, park2021nerfies}, have made impressive results on rendering photo-realistic images for the view-synthesis problems. 

% These works model the scene as a radiance field of color predicted by a neural network, making the representation an implicit function. The rendering process is an integration of the radiance along a ray.

Another line of work focuses on differentiable rasterization~\cite{kato2018neural, loper2014opendr, vicini2022differentiable, liu2019soft, li2020differentiable, ravi2020accelerating}. These works aim to replace the traditional rasterization process in computer graphics based on 2D projections of primitives such as polygons with z-buffering, with a differentiable rasterization process. 

% Representatively, \cite{liu2019soft} replaces the binary ray-triangle intersection function with z-buffering, neither of which is differentiable, with a soft aggregation function of pairwise influence from each triangle in a mesh to each pixel in the image, which are both differentiable. Through this modification, complex 3D geometry such as a human body can be estimated from multiple viewpoints of images.

While differentiable, these methods are limited by the intrinsic difficulty of modeling single or multiple bounces of light in a scene, which can be modeled with physics-based differentiable ray tracing \cite{li2018differentiable, zhang2019differential, nimier2019mitsuba, hu2019difftaichi, vicini2022differentiable, jiang2020sdfdiff}. In our problem, because humans are light sources and we need to perform differentiable rendering of one-bounce reflection, we extended Soft Rasterizer~\cite{liu2019soft}.

\textbf{Single-View 3D Reconstruction.}
From a practical point of view, obtaining 3D ground truth for supervision is often difficult and expensive~\cite{h36m_pami}. In terms of the quantity of data available, the unlabeled 3D data is not comparable to the 2D data on the internet. This spurs a long-standing interest from the general computer vision community to pursue 3D reconstruction with as little information as a single-view\cite{cmrKanazawa18,liu2019soft,ucmrGoel20,wu2020unsupervised,li2020self,ye2021shelf, liu2022shadows}.

In addition to general 3D object reconstruction, another line of research focus on the 3D reconstruction of human body from single-view images and videos \cite{lin2021end, kocabas2020vibe, kolotouros2019learning, moon2020i2l, rempe2021humor, pavlakos2019expressive}. Representatively, SMPL-X \cite{pavlakos2019expressive} is an expressive whole-body model with details around hands and faces, represented as a triangle mesh with 10,475 vertices. In the same paper, SMPLify-X was proposed to estimate an SMPL-X model from just a single RGB image. This is done by first detecting human keypoints from the image with an off-the-shelf keypoint detector~\cite{cao2017realtime, cao2017realtime, sun2019deep, cheng2020higherhrnet, fang2017rmpe, bazarevsky2020blazepose, li2020simple, guler2018densepose}. Then the parameters of an SMPL-X model is optimized to fit the keypoints which serve as the observation of human in the 2D image.

% Previously, \cite{maeda2019thermal} studied the problem of thermal non-line-of-sight imaging, proposing an image formation model and algorithms to perform object shape recovery and human detection from thermal reflection images. The primary differences from our work is two-fold: 1.we generalize the reflective surfaces to be non-planar, which is crucial for real-world application 2.our framework is based on 3D generative models and differentiable rendering of reflection to jointly estimate the shape of both 3D objects and 3D human.

\textbf{3D Generative Model.}
Our system utilizes generative models for both objects and humans. For 3D objects, generative models are usually trained with synthetic datasets composed of CAD models\cite{chang2015shapenet}. Different generative architectures including VAE \cite{guan2020generalized, brock2016generative, wu2019sagnet, guan2020generalized}, GAN \cite{ramasinghe2020spectral, ramasinghe2020spectral, hui2020progressive, wu2016learning}, normalizing flow \cite{kim2020softflow, klokov2020discrete}, and diffusion models \cite{luo2021diffusion} were proposed to generate objects in meshes, point clouds, or voxels. More recently, implicit 3D representation, or coordinate-based models, become a popular choice of modality to perform generative tasks \cite{hao2020dualsdf, mescheder2019occupancy, park2019deepsdf, deng2021deformed, ibing20213d, niemeyer2021giraffe}.

For humans, \cite{pavlakos2019expressive, tiwari2022pose} proposed generative models for 3D humans, represented as SMPL-X models. In \cite{pavlakos2019expressive}, a VAE is trained to generate human poses from 4 datasets including Human3.6M, LSP, CMU Panoptic, and PosePriors \cite{yu2016deep, han2017space, nie2019single, akhter2015pose}. The VAE samples a latent vector from a high-dimensional Gaussian distribution and generates a human pose vector. This pose vector is applied with a sparse linear regressor and a linear blend skinning function to generate a triangle mesh in a fully differentiable manner.

% This pose vector describing a 3D human skeleton is then applied with a sparse linear regressor to generate vertices to generate triangle meshes with a linear blend skinning function. Through this fully differentiable process, gradients of the coordinates of each vertex on a human body w.r.t. the input latent vector to the pose VAE, can be calculated for optimization tasks.

\textbf{Thermal Computer Vision.}
Previous work has applied computer vision to thermal images for various problems \cite{kutuk2022semantic, treptow2006real, gan2022unsupervised, heo2004fusion, davis2005two, chen2020multi, rivadeneira2022thermal}. ContactDB \cite{brahmbhatt2019contactdb} used thermal imaging to obtain accurate human grasps of everyday objects for robotics applications. \cite{maeda2019thermal} studied the problem of thermal non-line-of-sight imaging. In comparison, this work focuses on the 3D reconstruction of people from their thermal reflections in non-planar objects. Other work pursues 3D reconstruction of objects from thermal images \cite{chen20153d, schramm2022combining, maset2017photogrammetric, sage20203d}. To our knowledge, we are the first to perform 3D reconstruction of humans from their thermal reflection. 

\section{Methods} \label{method}
Our system takes an RGBD image and a thermal image of everyday objects with thermally reflective surfaces and performs a 2-stage optimization to estimate 3D objects and a human not in sight from both cameras' perspectives. In the first stage, a 6 DoF pose, scale, and a neural signed distance function \cite{park2019deepsdf} are jointly estimated for each object present in the scene. In the second stage, the location, orientation, and pose of the human are jointly estimated to reconstruct the observed thermal reflection.

Section \ref{method:formation} formulates the problem we aim to solve. Section \ref{method:overview} gives an overview of the approach. Section \ref{method:generative} describes the generative models we used in our approach in detail. Section \ref{method:differentiable} lays out a differentiable rendering algorithm of human thermal reflection. Section \ref{method:optimization} formulates the optimization process and the objective functions.

\subsection{Problem Formulation} \label{method:formation}
% Similar to a normal RGB camera, a thermal camera measures the intensity of electromagnetic waves received at a pixel location. However, the wavelength of light measured by a thermal camera is much longer than that of a normal RGB camera. We chose to use a long-wavelength infrared (LWIR) camera with a spectral range of 8$\mu m$ - 14 $\mu m$ which covers the peak of radiation emitted by a human body \cite{humanradiation}.

We decompose a scene into 3 components: a human body, objects with specular surfaces in LWIR spectrum, and environmental heat sources. We first obtain a segmentation mask from each object in the scene from the RGBD image. To obtain the thermal reflection image, we perform ray tracing starting from the camera sensor to the light source -- the human body, under Helmholtz reciprocity. Assuming a pinhole camera model, let $\textbf{\n}$ be the surface normal of the object at point $\textbf{\p}$, $\textbf{\r}$ be the vector from the camera sensor to $\textbf{\p}$ and $\textbf{\r}'$ the reflected ray vector. We model the intensity of each pixel \textbf{$I_\x$} in the thermal camera as a binary value:
\begin{equation}
    I_\x = \begin{cases}  1, & \textbf{\r}' \textrm{ intersects with } \mathcal{T}_{\phi, T}(M_h) \\
    0, & \textrm{otherwise} \end{cases}
    \label{eq:thermal_binary}
\end{equation}
where $M_h$ represents the human shape in the form of a triangle mesh, and $\mathcal{T}_{\phi, T}$ represents an SE(3) transformation matrix parameterized by rotation, translation, and scale. With background subtraction, the noise coming from environmental heat sources can be mitigated.

As described in figure \ref{fig:teaser}, the calibrated thermal camera and RGBD camera with known intrinsic matrix and unknown extrinsic matrix capture an RGB image, a depth map, and a thermal image. Given these images as our observation containing N objects, we solve for the following 7 variables via optimization: locations $\{\T_{obj}\}_{i=0}^N$, rotations $\{\phi_{obj}\}_{i=0}^N$, scales $\{s_{obj}\}_{i=0}^N$, and the shape $\{M_{obj}\}_{i=0}^N$ of the objects, location $\T_{h}$, rotation $\phi_{h}$, and the shape $M_{h}$ of the human, all in camera's perspective.

\begin{figure}
    \centering
    \includegraphics[width=\linewidth]{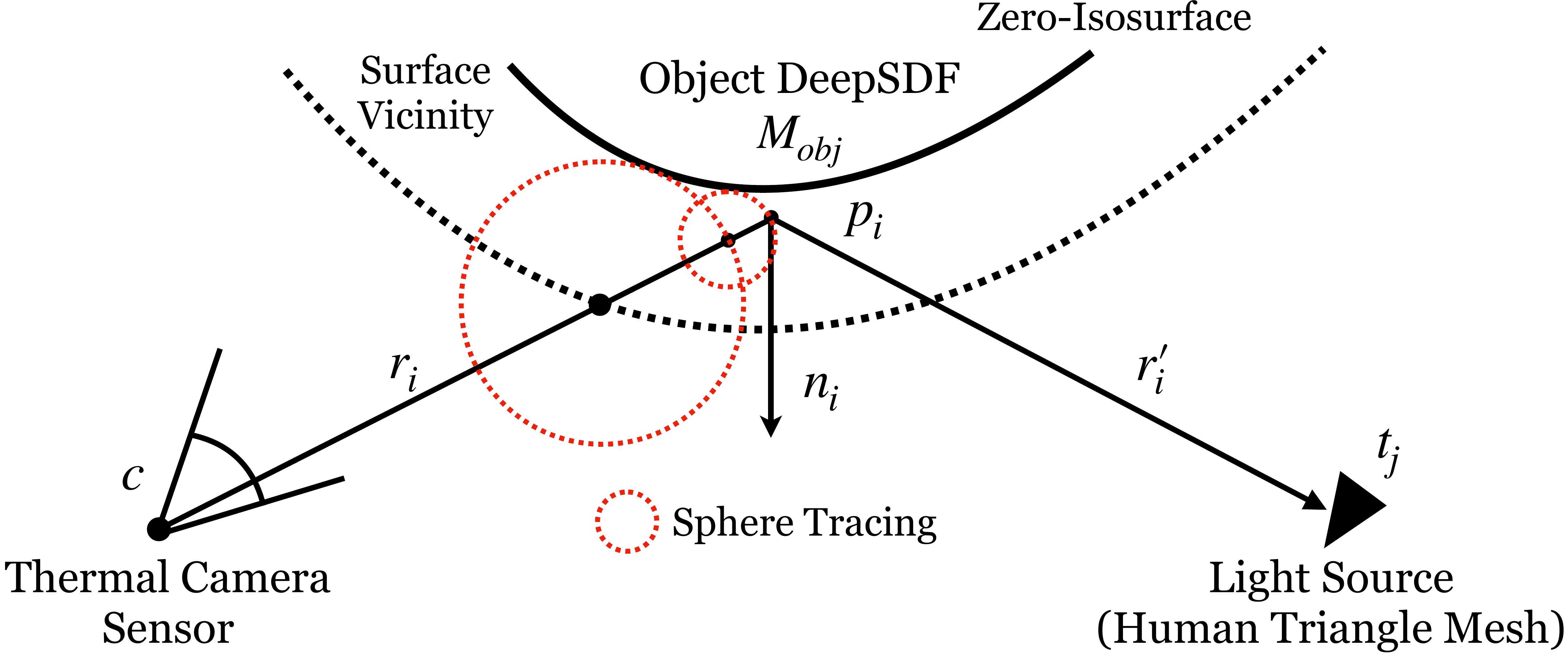}
    \caption{Differentiable Rendering of Reflection. Ray direction shown reverses the physical propagation direction of light by Helmholtz reciprocity.}
    \label{fig:ray-tracing}
\end{figure}

\subsection{Overview of Approach} \label{method:overview}
The optimization problem we are solving is severely under-constrained, so we choose to leverage the priors provided by pretrained generative models. As described in figure \ref{fig:method}, we first randomly sample the aforementioned 7 variables as initial input to the generative models to generate a 3D human and objects in the scene. Then we perform a differentiable rendering of human thermal reflection. For every ray from the camera sensor that intersects with an object, we can analytically calculate the reflected ray vector, given that the surface normals of the objects are defined by the output of the object generative model. With these reflected ray vectors, we can render a binary reflection image based on whether the reflected ray vectors intersect with humans, whose exact 3D shape and location are defined by the output of the human generative model. The optimization objective is to maximize the similarity between the rendered reflection image and the observed image captured by the thermal camera.

In order for such a pipeline to be differentiable, we need both the generative models of humans and objects, as well as the rendering algorithm, to be differentiable. In the following sections, we will describe how we achieve this.

\subsection{Generative Models} \label{method:generative}
\textbf{Object: DeepSDF.} 
We decided to use DeepSDF \cite{park2019deepsdf} as our generative models for objects. SDF, or signed distance function, is a function between a point in space and its orthogonal distance to the closest surface. In essence, DeepSDF is an SDF parameterized by a neural network $G_{obj}$ whose input is a 3D coordinate $\p$ and output is a signed distance $s$. Following \cite{park2019deepsdf}, we condition a DeepSDF model on a latent vector $\z_{obj}$ from a probabilistic latent space to make them generative model:
\begin{equation}
    G_{obj}(\p, \z_{obj}) = s : \p \in \mathbb{R}^3, s \in \mathbb{R}
\end{equation}

\begin{figure}[t]
    \centering
    \vspace{-0.2cm}
    \includegraphics[width=\linewidth]{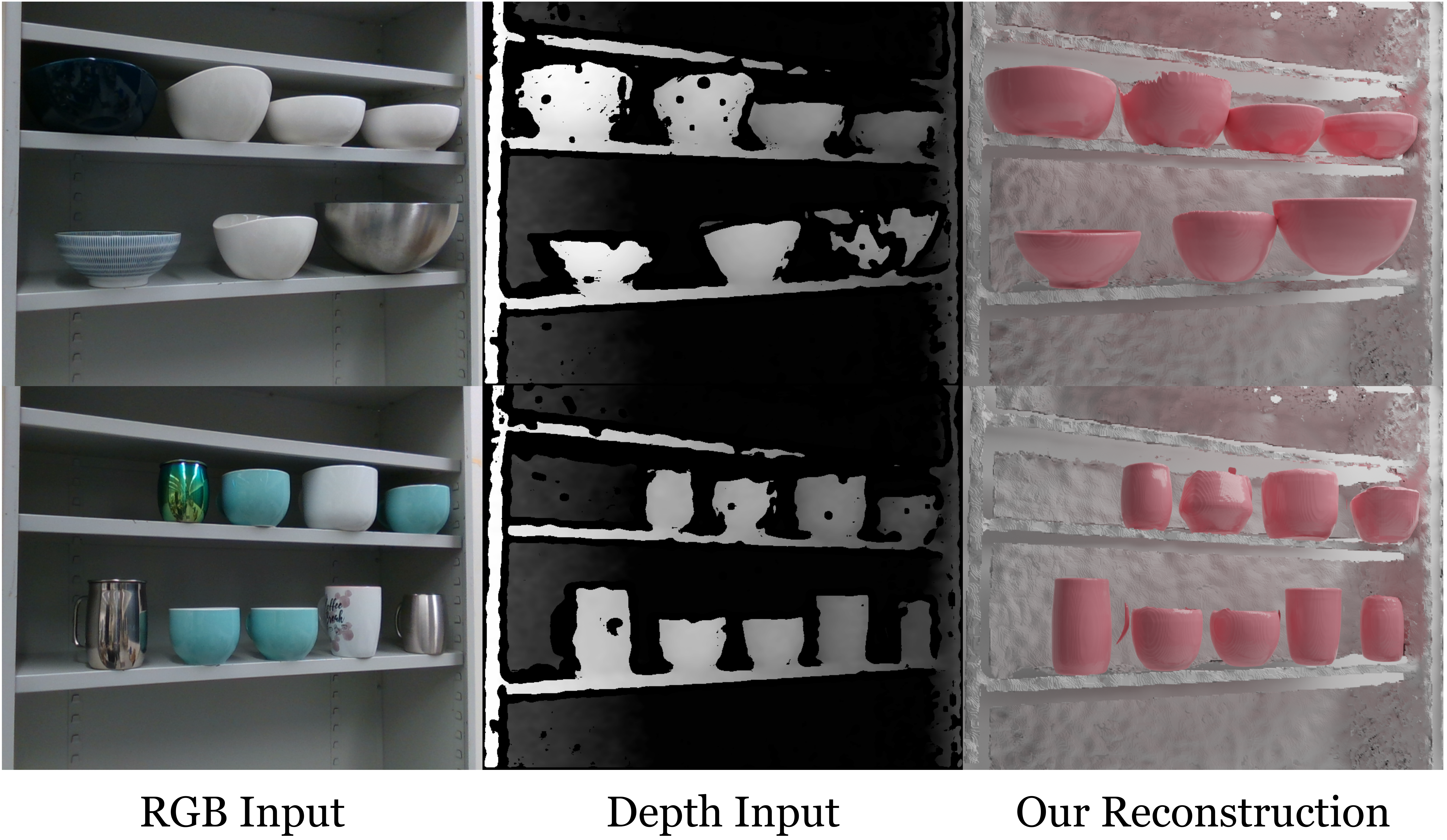}
    \caption{3D Object Reconstruction from RGBD}
    \label{fig:object}
\end{figure}

\textbf{Human: SMPL-X.}
We adopted SMPL-X~\cite{pavlakos2019expressive} as our generative models of 3D humans. Broadly, SMPL-X is composed of 2 components. The first is a variational autoencoder (VAE) that projects a latent vector $\z_h$ sampled from a probabilistic latent space with Gaussin prior to the human pose space, in the form of rotations of human body joints. The generated human body pose is then applied with a differentiable sparse linear regressor to generate vertices and triangle meshes representing the surface skins of a human body. Because both the VAE and the linear vertex regressor are differentiable, the location of each vertex is differentiable w.r.t. the latent vector $\z_h$.

\begin{figure*}[t]
    \centering
    \includegraphics[width=\linewidth]{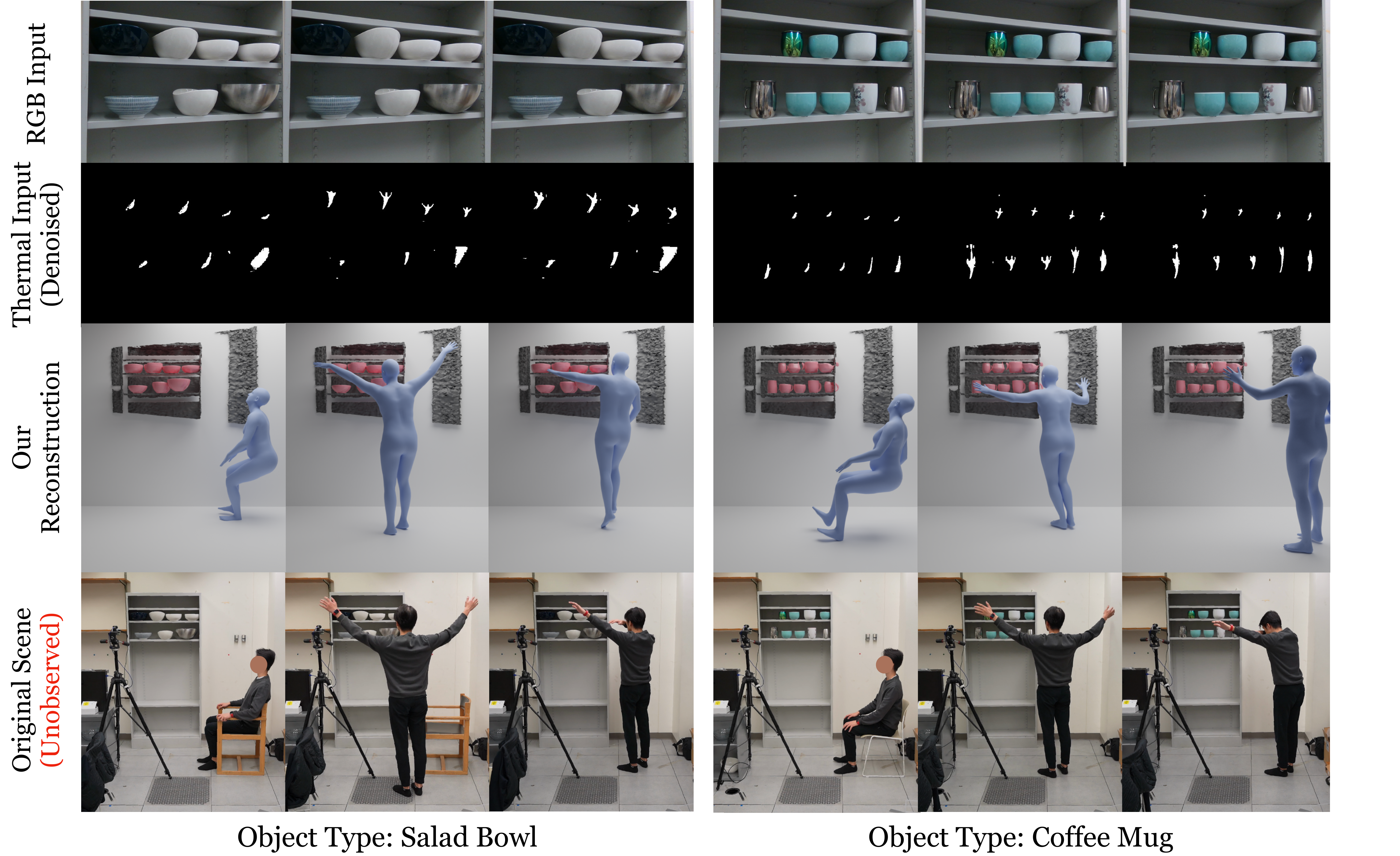}
    \caption{3D Human Reconstruction (visualized from another camera view). From an RGBD image, we recover the 3D location $\T_{obj}$, pose $\phi_{obj}$, and shape $\z_{obj}$ of each object. A marching cube visualization of the reconstructed 3D objects is shown in \textcolor{object}{pink}. With reconstructed objects, we recover 3D location $\T_{h}$, pose $\phi_{h}$, and shape $\z_{h}$ of the human from a denoised thermal input showing reflections of the human on object surfaces, which we visualize in \textcolor{human}{blue}. We also include the original scene and our reconstruction from a calibrated third-camera view for comparison. This image is \textbf{not seen} by our system during reconstruction. The black mesh where the objects are located is the depth pointclouds captured by the RGBD camera.}
    \label{fig:human}
\end{figure*}

\subsection{Differentiable Rendering of Reflection} \label{method:differentiable}
The information we have from the thermal image of objects is a reflected human silhouette. Soft rasterizer (SoftRas) \cite{liu2019soft} is a method of choice to perform differentiable rendering from 2D silhouette images. However, SoftRas is a differentiable rasterization algorithm, which does not directly apply to reflection, especially when the reflective surface is a curved surface defined by a DeepSDF. To overcome this limitation, we extended SoftRas to ray tracing under non-planar reflection off the zero-isosurface  of a DeepSDF. This process is visualized in figure \ref{fig:ray-tracing}.

\textbf{DeepSDF Depth Estimation.}
The complex geometry of an everyday object prevents us from projecting all triangles to the 2D image plane as in \cite{liu2019soft}. Thus, we need to march rays $\{\r_i\}$ from camera sensor $\c$, through the reflection point on the surface $\{\p_i\}$ with a surface normal $\{\n_i\}$, to the reflected rays $\{\r_i'\}$. To obtain the intersection point with the surface $\{\p_i\}$ given an SDF representation of an object, we need a differentiable method to extract the zero-isosurface and calculate the depth of the surface along the incoming ray $\r_i$. Previously, \cite{zakharov2020autolabeling} proposed to perform surface projection by first grid-searching for a point close to the zero-isosurface, then projecting along gradient direction $\frac{\partial G}{\partial\p}$ with the predicted distance. However, because the gradient direction is not in the same direction as the incoming ray, performing such an operation could yield a point far from the intersection point between the incoming ray and zero-isosurface, especially when the attack angle is small. To mitigate this error, we perform finite steps of sphere tracing along the ray to estimate the intersection point $\{\p_i\}$ as shown in figure \ref{fig:ray-tracing}.

\textbf{DeepSDF Surface Normal.}
With the estimated intersection point $\{\p_i\}$ between $\{\r_i\}$ and the surface of the object, we calculate the surface normal of the object at $\{\p_i\}$:
\begin{equation}
    \n_i = \frac{\partial G_{obj}(\p_i, \z_{obj})}{\partial \p_i}
\end{equation}
We can then calculate the reflected ray vector as:
\begin{equation}
    \r_i' =  \r_i + 2 \cdot \r_i \cdot \frac{\n_i}{\|\n_i\| _2}
\end{equation}

\begin{figure*}[t]
    \begin{center}
    \includegraphics[width=1\textwidth]{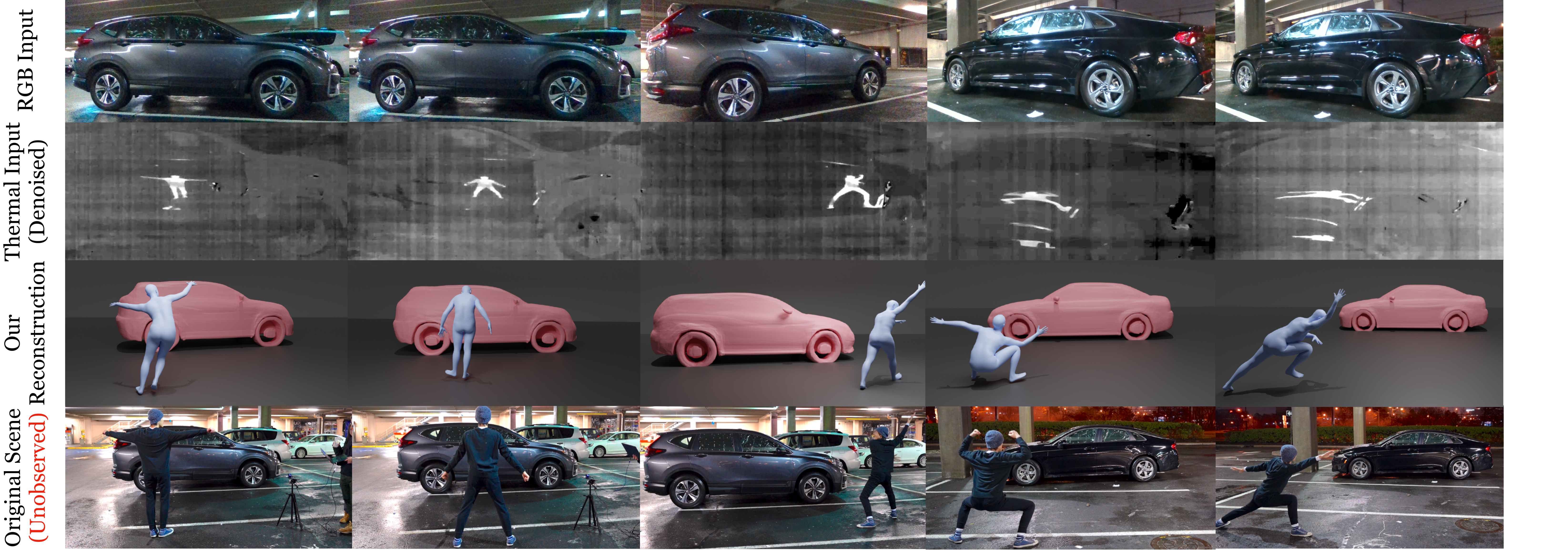}
    \captionof{figure}{
    Real-world 3D human reconstruction from thermal reflections of cars. A diverse set of human poses can be reconstructed by using the surfaces of different types of cars as infrared mirrors. RGB input (1st row) and thermal input (2nd row) captured by a depth camera and a thermal camera are used as input to our method. Our reconstruction (3rd row) is compared with the original scene (4th row), both rendered/captured from another camera viewpoint.
    }
    \vspace{-1em}
    \label{fig:car}
    \end{center}
\end{figure*}

\textbf{3D Ray-Triangle Distance.}  We can then calculate the pairwise distance matrix, denoted as $\mathcal{D}_{i, j}$ between each reflected ray $\r_i'$ and each triangle $t_j \in \{\M_h\}$, where $\M_h$ represents human body mesh. Each element in the distance matrix $d_{i, j} \in \mathcal{D}$ can be expressed as a differentiable function of vertices of $t_j$ and the reflected ray vector $\r_i'$. We can also obtain a ray-triangle intersection matrix $\Lambda$ with the same dimension as the distance matrix. Since the value of the ray-triangle intersection is binary, this calculation is not required to be differentiable.

\textbf{Differentiable Ray Occupancy.} Following SoftRas~\cite{liu2019soft}, we define the influence of each triangle $t_j$ on each ray $r_i'$ where the influence is expressed as a function of distance $d_{i, j}$:
\begin{equation}
    d'_{i, j} = \textrm{sigmoid}\left(\lambda_{i, j} \frac{d^2_{i, j}}{\sigma}\right), \;\; \lambda_{i, j} \in \Lambda, \;\; d_{i, j} \in \mathcal{D}
\end{equation}
where $\lambda_{i, j} = 1$ if reflected ray $\r_i'$ intersects with triangle $\t_j$, otherwise $-1$. $d_{i, j}$ denotes the distance between ray $\r_i'$ and triangle $\t_j$, $\sigma$ is a hyperparameter that controls the ``softness'' of the influence. We then aggregate the influence of each triangle for a ray reflected $\r_i'$ to obtain the estimated binary occupancy of the ray by human body mesh $\M_h$:
\begin{equation}
    \hat{I_i} = \mathcal{A}(\{\mathcal{D}\}_j) = 1 - \Pi_j(1 - d_{i, j})
\end{equation}
The estimated binary occupancy of ray $\hat{I_i}$ is a value between 0 and 1 and is compared with the ground truth binary thermal image defined in Eq. \ref{eq:thermal_binary}.

\subsection{Optimization for Inference} \label{method:optimization}
\textbf{3D Object Reconstruction.} We first estimate the 6 DoF pose, scale, and shape of the objects present in the scene following a similar method as in \cite{irshad2022shapo}. We optimize the locations $\{\T_{obj}\}_{i=0}^N$, rotations $\{\phi_{obj}\}_{i=0}^N$, scale $\{\s_{obj}\}_{i=0}^N$, and the shape of the objects $\{\z_{obj}\}_{i=0}^N$, where $\{\z_{obj}\}_{i=0}^N$ are latent variables sampled from the probabilistic latent space of DeepSDF $\G_{obj}$ s.t. $\M_{obj} = G_{obj}(\z_{obj})$. For each object, we minimize the objective:
\begin{equation}
    \mathcal{L}_{obj} = \mathcal{L}_{depth} + \mathcal{L}_{mask} + \mathcal{L}_{prior}
\end{equation}
where $\mathcal{L}_{depth}$ is the $L_1$ loss between the estimated depth map and the measured depth map, $\mathcal{L}_{mask}$ denotes a pixel-wise $L_2$ loss between the estimated segmentation mask and the observed segmentation mask obtained from RGB observation, and $\mathcal{L}_{prior}$ is a shape prior regularization term.\looseness-1

\begin{figure*}[t]
    \centering
    \includegraphics[width=\linewidth]{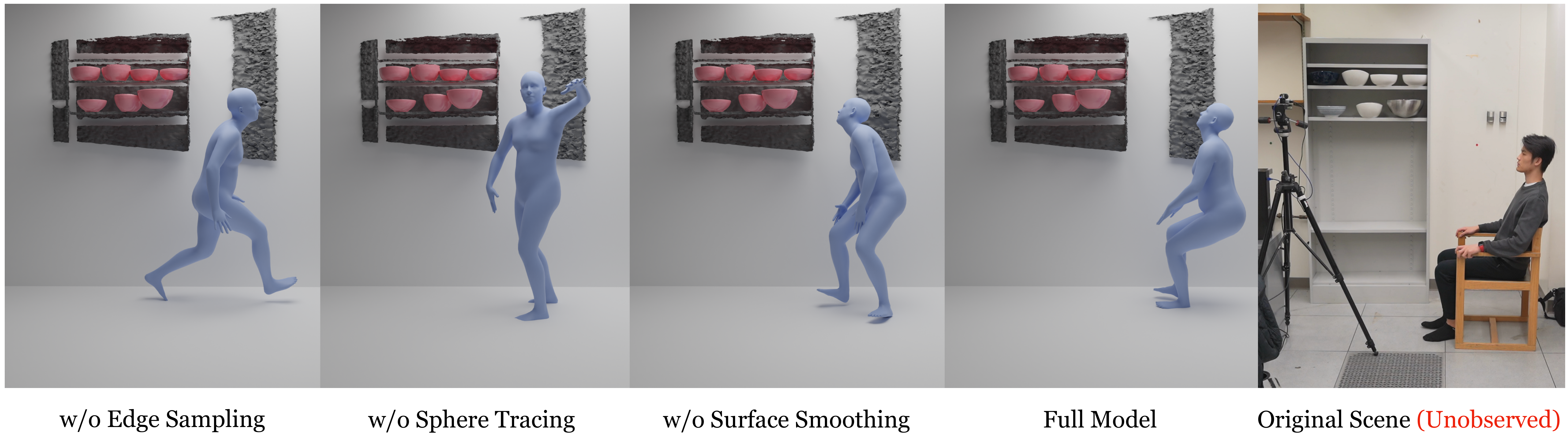}
    \caption{Visualization of reconstruction obtained from ablated variations of our full model. While all variations can still find the 3D location of humans relatively accurately, the fine-grained details of human poses are significantly improved in our full model.}
    \vspace{-0.2cm}
    \label{fig:ablation}
\end{figure*}

\textbf{3D Human Reconstruction} Given the estimated translations, rotations, scales, and shape latent vectors from 3D object reconstruction, we optimize translation $\T_{h}$, rotation $\phi_{h}$, and shape $\z_{h}$ of the human where $\z_{h}$ is the latent vector sampled from the pose VAE in SMPL-X s.t. $\M_{h} = G_{h}(\z_{h})$. Upon obtaining the estimated reflection image $\hat{\I}$ and observed thermal silhouette image $\I$, we minimize the objective:
\begin{equation}
    \mathcal{L}_{human} = \mathcal{L}_{silhouette} + \mathcal{L}_{prior}
\end{equation}
where 
\begin{equation}
    \mathcal{L}_{silhouette} = 1 - \frac{\|\hat{\I} \otimes \I\|_1}{\|\hat{\I} \oplus \I - \hat{\I} \otimes \I\|_1}
\end{equation}
and $\mathcal{L}_{prior}$ is an $L_2$ regularization term on the human latent vector $\z_h$

\section{Experiments} \label{experiments}
The goal of our experiments is to validate our hypothesis that LWIR thermal reflection on everyday objects provides sufficient information to perform accurate 3D human reconstruction in the real world. In section \ref{experiment:obj}, we first demonstrate the accurate 3D reconstruction of objects from a single RGBD image, which serves as a foundation for 3D human reconstruction from reflection. We showcase our results on 3D human reconstruction with different poses and object types with everyday objects (section~\ref{experiment:human}) and cars (section ~\ref{experiment:car}). Lastly, in section \ref{experiment:ablation} we perform quantitative and qualitative ablation studies to evaluate the effectiveness of our technical approach.

\subsection{3D Object Reconstruction} \label{experiment:obj}
Real-world depth sensors are subject to often significant measurement errors and are sensitive to lighting conditions (assuming an active stereo sensor). The surface depth estimated is often noisy, non-smooth, and full of ``holes'', as shown in figure \ref{fig:object}. Performing differentiable rendering of reflection using the direct output of the depth sensor will necessarily introduce an excessive amount of noise, given the reflected ray direction is calculated from the surface normal. Therefore, we opted to perform 3D object reconstruction from RGBD input first, then use the reconstructed surfaces for differentiable ray tracing.

In figure \ref{fig:object}, we visualize the reconstructed objects from the RGBD input. Because our 3D representation of objects is an implicit function -- DeepSDF, we perform marching cubes to extract the zero-isosurface of each object generated from the latent vector $\z_{obj}$. We then applied the SE(3) transformation matrix which is calculated from the estimated location $T_{obj}$ and pose $\phi_{obj}$.

\begin{table*}
    \setlength{\tabcolsep}{4pt}
    \centering
    \begin{tabular}{M{2.8cm} M{1.2cm} M{1.8cm} M{1.8cm} M{1.8cm} M{1.8cm} M{1.8cm}}
        \toprule
        Evaluation Method & Object Type & w/o Edge Sampling & w/o Sphere Tracing & w/o Surface Smoothing & \cellcolor{Gray}{Full Model} & Random  \\
        \midrule
        2D Keypoints \cite{fang2017rmpe} & Bowl & 0.231 & 0.224 & 0.145 & \cellcolor{Gray}{\textbf{0.116}} & 0.346  \\
        2D Keypoints \cite{fang2017rmpe} & Mug & 0.101 & 0.209 & 0.109 & \cellcolor{Gray}{\textbf{0.094}} & 0.371 \\
        3D Skeleton \cite{li2021hybrik} & Bowl & 0.309 & 0.272 & 0.212 & \cellcolor{Gray}{\textbf{0.152}} & 0.322 \\
        3D Skeleton \cite{li2021hybrik} & Mug & 0.223 & 0.215 & 0.202 & \cellcolor{Gray}{\textbf{0.126}} & 0.317 \\
        \bottomrule
    \end{tabular}
    \caption{Quantitative evaluation of our reconstructed 3D human. We used two evaluation methods by comparing the extracted 2D keypoints and 3D skeleton with a calibrated 3rd camera view. Object type indicates the type of objects serving as reflectors. Columns 3-5 are three variations of our full model with some parts ablated. Random shows the corresponding metric if a random sample were to be drawn from the HumanEva\cite{sigal2010humaneva} dataset, which includes diverse poses in daily human activities. Numbers show the average normalized Euclidean distance between reconstruction and ground truth.}
    \vspace{-0.2cm}
    \label{tab:ablation}
\end{table*}

As shown in figure \ref{fig:object}, the location, pose, and shape of 3D objects can be faithfully reconstructed. Most importantly, we are able to obtain a high-fidelity, smooth, and accurate object surface without an explicit regularization on surface smoothness, which sets the foundation for the differentiable rendering of reflection. The successful reconstruction even when the depth input is noisy can be largely attributed to the object priors provided by searching in the latent space of a pretrained generative model. Generative priors such as a bowl is usually symmetric, the outside surface of a mug is often smooth, are enforced during the optimization process.

\subsection{3D Human Reconstruction} \label{experiment:human}

Given the reconstructed objects represented as individual DeepSDF models and their locations, we perform joint optimization of human location $\T_{h}$, orientation $\phi_{h}$, and shape $\z_h$. The input to the differentiable rendering algorithm is a single binary thermal reflection image, representing a mask of human silhouette on each reflective object, as shown in figure~\ref{fig:human}. The binary mask of reflection is obtained from the thermal camera pointing towards the reflective objects, with simple denoising and thresholding. 
% We record a video of thermal images, perform background subtraction to mitigate the influence of environmental heat sources, denoise the image with erosion and dilation convolution filters, then perform thresholding.
In addition to RGBD and thermal cameras, we put a third calibrated camera in the scene to capture the scene from another angle for evaluation and visualization. Note that any images from this camera are not used as input to our system.

We render the reconstruction from the third camera's perspective for comparison with the original scene at the exact time input data was captured. As shown in figure \ref{fig:human}, the output of our method very accurately reconstructs the original scene. Note that the subject in the original scene is wearing normal clothing and the data is collected in a normal office environment without special lab environmental control. Besides, the objects used to reflect human thermal radiation are everyday objects with a variety of textures and materials that we purchased from supermarkets. This indicates the robustness of our system and its practical applicability to various settings.

\subsection{Cars as Infrared Mirrors} \label{experiment:car}
Non-line-of-sight information of human activity plays a crucial role in the safe deployment of autonomous driving systems. Therefore, we showcase an experiment where we use cars as infrared mirrors to reconstruct the 3D location, orientation, and shape of a pedestrian that's not in the line-of-site of a camera system. In figure~\ref{fig:car}, we show the results in a similar fashion as figure~\ref{fig:human}. 3D reconstruction from thermal imaging could allow new opportunities for autonomous vehicles to sense and safely avoid occluded pedestrians.

% We believe our framework could lead to important assistive technology in modern autonomous driving systems.

\subsection{Ablation Studies} \label{experiment:ablation}
To solve the extremely under-constrained and challenging problem, we made a lot of design decisions that turned out to be crucial to the quality of reconstruction. To evaluate the effectiveness of our technical approach, we perform ablation studies and compare our reconstruction with a baseline.  We have included both quantitative evaluations as well as qualitative visualizations. Here we described some representative design decisions in detail.

\textbf{Edge Sampling.}
As pointed out by \cite{li2018differentiable}, edge sampling plays an important role in differentiable ray tracing. This is even more significant for human reflection silhouettes. In addition, unless a person is standing right in front of the reflector, the reflection silhouette usually occupies a small region of the thermal image. We therefore perform edge detection on the reflection image to extract edges of human silhouette and sampling ray with a probability distribution concentrated at the vicinity of these edges and increasing the concentration as training progresses as a type of curriculum training.

\textbf{Sphere Tracing.}
As we've described in \ref{method:differentiable}, direct surface projection from the vicinity of an SDF will yield a point far from the real intersection between the incoming ray and the zero-isosurface of the SDF. Therefore, we perform 3 steps of sphere tracing to estimate the intersection point on the object.

\begin{figure}[t]
    \centering
    \includegraphics[width=0.95\linewidth]{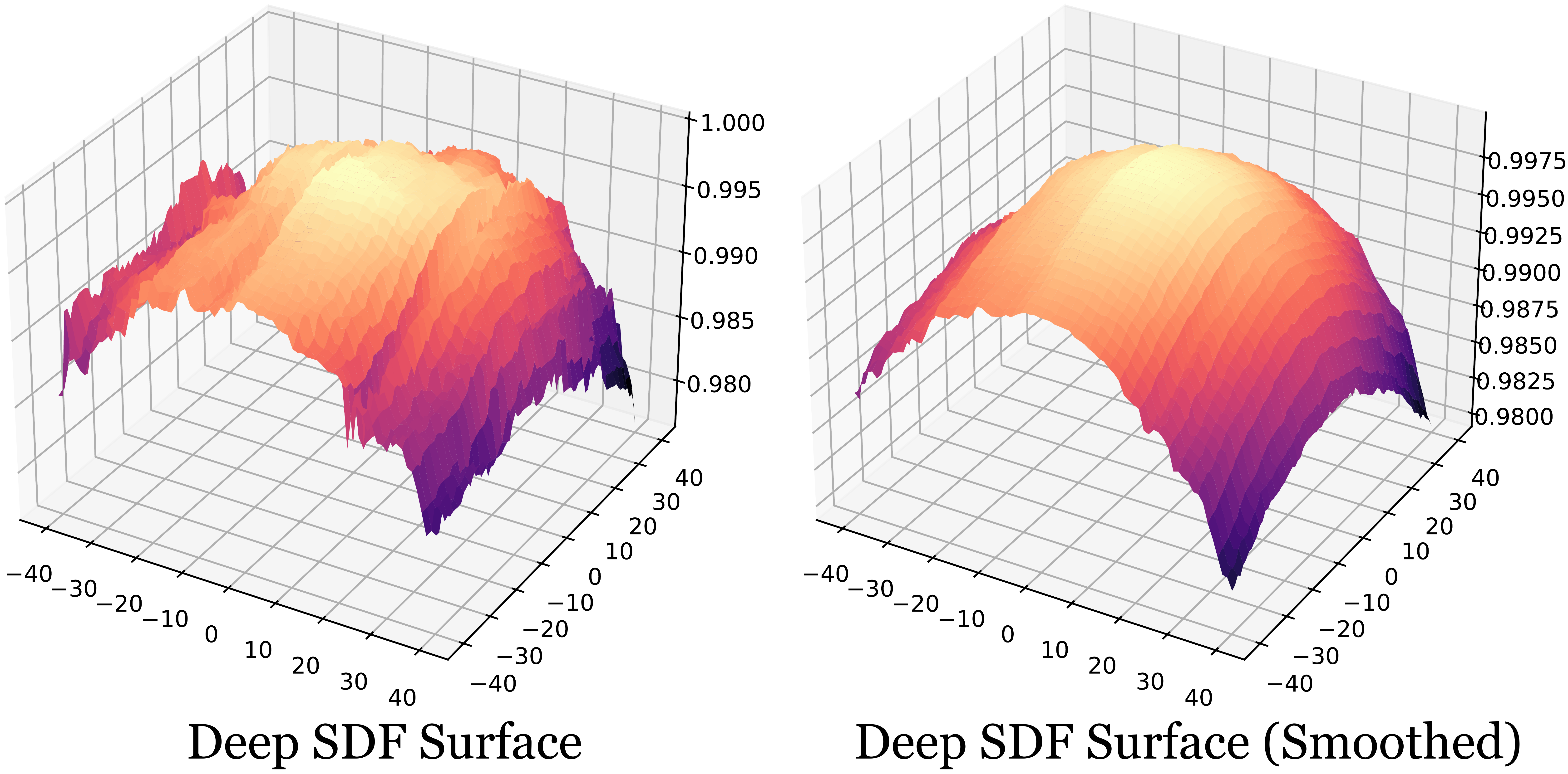}
    \caption{Visualization of DeepSDF Surface Normals within a 1.6 cm$\times$1.6 cm area. From the visualization, we can clearly see an improvement in surface smoothness at a small scale, which is beneficial to the differentiable rendering process. The X-Y plane (horizontal) indicates the location on a surface with a step size of 0.2mm. Given the unit surface normal vector at a point on the grid $(x, y)$, we compute its dot product with the unit surface normal vector at $(0, 0)$, and plot this value on the Z-axis. This shows the curvature of the surface as well as its level of smoothness.}
    \vspace{-0.2cm}
    \label{fig:surface}
\end{figure}

\textbf{Surface Smoothing.}
From experiments, we discovered that even after we perform sphere tracing, the reflection surface normals are still noisy, causing the differentiable rendering algorithm to produce a noisy reflection. This effectively injects noise into the gradients, making the optimization more challenging. We discovered that this is caused by the reconstructed DeepSDF having a locally non-smooth zero iso-surface. In figure \ref{fig:surface}, we visualize the surface normals calculated from a small region of zero-isosurface which shows the non-smoothness. To mitigate this error, we perform surface smoothing during differentiable rendering by sampling 8 neighboring rays surrounding the main ray and averaging all estimated surface normals for reflection calculation.

\textbf{Evaluation.}
We evaluate our reconstruction as well as the 3 aforementioned ablated methods by comparing the 2D keypoints and 3D skeleton estimated from synchronized images captured by a calibrated third camera. We used \cite{fang2017rmpe} for 2D keypoints detection and \cite{li2021hybrik} for 3D skeleton estimation. For comparison, we compared the reconstruction to 200 randomly sampled 2D human keypoints and 3D skeletons from the HumanEva dataset \cite{sigal2010humaneva}.

Both the quantitative experiments and qualitative visualizations have shown the effectiveness of our technical approach as well as the design decisions. Particularly, we believe our findings regarding differentiable rendering of reflections on implicit surfaces will provide insights to other computer vision researchers working with reflections.

\section{Conclusion}
This paper shows that 3D position and pose of a human can be reconstructed from a single thermal image of everyday objects reflecting human thermal radiations. We approach this problem by combining the priors learned by pretrained 3D generative models and differentiable rendering of reflections. By formulating the problem as an optimization problem, we perform analysis by synthesis to explain the observations. We believe thermal cameras are powerful tools to study human activities in daily environments and integrating them with modern computer vision models will bring out many downstream applications in robotics, graphics, and 3D perception.

{\small\textbf{Acknowledgements:} This research is based on work partially supported by the Toyota Research Institute, the NSF NRI Award \#1925157, and the NSF CAREER Award \#2046910. We acknowledge Shree Nayar, Shuran Song, Runlin Xu, James Tompkin, Mark Sheinin, Mia Chiquier, Jeremy Klotz for helpful feedback, and Su Li, Dylan Chen, Sophia Su for helping with data collection.}

%%%%%%%%% REFERENCES
{\small
\bibliographystyle{ieee_fullname}
\bibliography{thermal}

\begin{thebibliography}{10}\itemsep=-1pt

\bibitem{akhter2015pose}
Ijaz Akhter and Michael~J Black.
\newblock Pose-conditioned joint angle limits for 3d human pose reconstruction.
\newblock In {\em Proceedings of the IEEE conference on computer vision and
  pattern recognition}, pages 1446--1455, 2015.

\bibitem{barron2021mip}
Jonathan~T Barron, Ben Mildenhall, Matthew Tancik, Peter Hedman, Ricardo
  Martin-Brualla, and Pratul~P Srinivasan.
\newblock Mip-nerf: A multiscale representation for anti-aliasing neural
  radiance fields.
\newblock In {\em Proceedings of the IEEE/CVF International Conference on
  Computer Vision}, pages 5855--5864, 2021.

\bibitem{barron2022mip}
Jonathan~T Barron, Ben Mildenhall, Dor Verbin, Pratul~P Srinivasan, and Peter
  Hedman.
\newblock Mip-nerf 360: Unbounded anti-aliased neural radiance fields.
\newblock In {\em Proceedings of the IEEE/CVF Conference on Computer Vision and
  Pattern Recognition}, pages 5470--5479, 2022.

\bibitem{batchuluun2020region}
Ganbayar Batchuluun, Na~Rae Baek, Dat~Tien Nguyen, Tuyen~Danh Pham, and
  Kang~Ryoung Park.
\newblock Region-based removal of thermal reflection using pruned fully
  convolutional network.
\newblock {\em IEEE Access}, 8:75741--75760, 2020.

\bibitem{batchuluun2019study}
Ganbayar Batchuluun, Hyo~Sik Yoon, Dat~Tien Nguyen, Tuyen~Danh Pham, and
  Kang~Ryoung Park.
\newblock A study on the elimination of thermal reflections.
\newblock {\em IEEE Access}, 7:174597--174611, 2019.

\bibitem{bazarevsky2020blazepose}
Valentin Bazarevsky, Ivan Grishchenko, Karthik Raveendran, Tyler Zhu, Fan
  Zhang, and Matthias Grundmann.
\newblock Blazepose: On-device real-time body pose tracking.
\newblock {\em arXiv preprint arXiv:2006.10204}, 2020.

\bibitem{bennett1961relation}
HE Bennett and JO119764 Porteus.
\newblock Relation between surface roughness and specular reflectance at normal
  incidence.
\newblock {\em JOSA}, 51(2):123--129, 1961.

\bibitem{brahmbhatt2019contactdb}
Samarth Brahmbhatt, Cusuh Ham, Charles~C Kemp, and James Hays.
\newblock Contactdb: Analyzing and predicting grasp contact via thermal
  imaging.
\newblock In {\em Proceedings of the IEEE/CVF conference on computer vision and
  pattern recognition}, pages 8709--8719, 2019.

\bibitem{brock2016generative}
Andrew Brock, Theodore Lim, James~M Ritchie, and Nick Weston.
\newblock Generative and discriminative voxel modeling with convolutional
  neural networks.
\newblock {\em arXiv preprint arXiv:1608.04236}, 2016.

\bibitem{cao2017realtime}
Zhe Cao, Tomas Simon, Shih-En Wei, and Yaser Sheikh.
\newblock Realtime multi-person 2d pose estimation using part affinity fields.
\newblock In {\em Proceedings of the IEEE conference on computer vision and
  pattern recognition}, pages 7291--7299, 2017.

\bibitem{chang2015shapenet}
Angel~X Chang, Thomas Funkhouser, Leonidas Guibas, Pat Hanrahan, Qixing Huang,
  Zimo Li, Silvio Savarese, Manolis Savva, Shuran Song, Hao Su, et~al.
\newblock Shapenet: An information-rich 3d model repository.
\newblock {\em arXiv preprint arXiv:1512.03012}, 2015.

\bibitem{chen20153d}
Chia-Yen Chen, Chia-Hung Yeh, Bao~Rong Chang, and Jun-Ming Pan.
\newblock 3d reconstruction from ir thermal images and reprojective
  evaluations.
\newblock {\em Mathematical Problems in Engineering}, 2015, 2015.

\bibitem{chen2020multi}
I-Chien Chen, Chang-Jen Wang, Chao-Kai Wen, and Shiow-Jyu Tzou.
\newblock Multi-person pose estimation using thermal images.
\newblock {\em IEEE Access}, 8:174964--174971, 2020.

\bibitem{cheng2020higherhrnet}
Bowen Cheng, Bin Xiao, Jingdong Wang, Honghui Shi, Thomas~S Huang, and Lei
  Zhang.
\newblock Higherhrnet: Scale-aware representation learning for bottom-up human
  pose estimation.
\newblock In {\em Proceedings of the IEEE/CVF conference on computer vision and
  pattern recognition}, pages 5386--5395, 2020.

\bibitem{davis2005two}
James~W Davis and Mark~A Keck.
\newblock A two-stage template approach to person detection in thermal imagery.
\newblock In {\em 2005 Seventh IEEE Workshops on Applications of Computer
  Vision (WACV/MOTION'05)-Volume 1}, volume~1, pages 364--369. IEEE, 2005.

\bibitem{deng2021deformed}
Yu Deng, Jiaolong Yang, and Xin Tong.
\newblock Deformed implicit field: Modeling 3d shapes with learned dense
  correspondence.
\newblock In {\em Proceedings of the IEEE/CVF Conference on Computer Vision and
  Pattern Recognition}, pages 10286--10296, 2021.

\bibitem{fang2017rmpe}
Hao-Shu Fang, Shuqin Xie, Yu-Wing Tai, and Cewu Lu.
\newblock Rmpe: Regional multi-person pose estimation.
\newblock In {\em Proceedings of the IEEE international conference on computer
  vision}, pages 2334--2343, 2017.

\bibitem{gan2022unsupervised}
Lu Gan, Connor Lee, and Soon-Jo Chung.
\newblock Unsupervised rgb-to-thermal domain adaptation via multi-domain
  attention network.
\newblock {\em arXiv preprint arXiv:2210.04367}, 2022.

\bibitem{ucmrGoel20}
Shubham Goel, Angjoo Kanazawa, , and Jitendra Malik.
\newblock Shape and viewpoints without keypoints.
\newblock In {\em ECCV}, 2020.

\bibitem{guan2020generalized}
Yanran Guan, Tansin Jahan, and Oliver van Kaick.
\newblock Generalized autoencoder for volumetric shape generation.
\newblock In {\em Proceedings of the IEEE/CVF Conference on Computer Vision and
  Pattern Recognition Workshops}, pages 268--269, 2020.

\bibitem{guler2018densepose}
R{\i}za~Alp G{\"u}ler, Natalia Neverova, and Iasonas Kokkinos.
\newblock Densepose: Dense human pose estimation in the wild.
\newblock In {\em Proceedings of the IEEE conference on computer vision and
  pattern recognition}, pages 7297--7306, 2018.

\bibitem{han2017space}
Fei Han, Brian Reily, William Hoff, and Hao Zhang.
\newblock Space-time representation of people based on 3d skeletal data: A
  review.
\newblock {\em Computer Vision and Image Understanding}, 158:85--105, 2017.

\bibitem{hao2020dualsdf}
Zekun Hao, Hadar Averbuch-Elor, Noah Snavely, and Serge Belongie.
\newblock Dualsdf: Semantic shape manipulation using a two-level
  representation.
\newblock In {\em Proceedings of the IEEE/CVF Conference on Computer Vision and
  Pattern Recognition}, pages 7631--7641, 2020.

\bibitem{heo2004fusion}
Jingu Heo, Seong~G Kong, Besma~R Abidi, and Mongi~A Abidi.
\newblock Fusion of visual and thermal signatures with eyeglass removal for
  robust face recognition.
\newblock In {\em 2004 Conference on Computer Vision and Pattern Recognition
  Workshop}, pages 122--122. IEEE, 2004.

\bibitem{hu2019difftaichi}
Yuanming Hu, Luke Anderson, Tzu-Mao Li, Qi Sun, Nathan Carr, Jonathan
  Ragan-Kelley, and Fr{\'e}do Durand.
\newblock Difftaichi: Differentiable programming for physical simulation.
\newblock {\em arXiv preprint arXiv:1910.00935}, 2019.

\bibitem{hui2020progressive}
Le Hui, Rui Xu, Jin Xie, Jianjun Qian, and Jian Yang.
\newblock Progressive point cloud deconvolution generation network.
\newblock In {\em European Conference on Computer Vision}, pages 397--413.
  Springer, 2020.

\bibitem{ibing20213d}
Moritz Ibing, Isaak Lim, and Leif Kobbelt.
\newblock 3d shape generation with grid-based implicit functions.
\newblock In {\em Proceedings of the IEEE/CVF Conference on Computer Vision and
  Pattern Recognition}, pages 13559--13568, 2021.

\bibitem{h36m_pami}
Catalin Ionescu, Dragos Papava, Vlad Olaru, and Cristian Sminchisescu.
\newblock Human3.6m: Large scale datasets and predictive methods for 3d human
  sensing in natural environments.
\newblock {\em IEEE Transactions on Pattern Analysis and Machine Intelligence},
  36(7):1325--1339, jul 2014.

\bibitem{irshad2022shapo}
Muhammad~Zubair Irshad, Sergey Zakharov, Rares Ambrus, Thomas Kollar, Zsolt
  Kira, and Adrien Gaidon.
\newblock Shapo: Implicit representations for multi-object shape, appearance,
  and pose optimization.
\newblock {\em arXiv preprint arXiv:2207.13691}, 2022.

\bibitem{jiang2020sdfdiff}
Yue Jiang, Dantong Ji, Zhizhong Han, and Matthias Zwicker.
\newblock Sdfdiff: Differentiable rendering of signed distance fields for 3d
  shape optimization.
\newblock In {\em Proceedings of the IEEE/CVF conference on computer vision and
  pattern recognition}, pages 1251--1261, 2020.

\bibitem{cmrKanazawa18}
Angjoo Kanazawa, Shubham Tulsiani, Alexei~A. Efros, and Jitendra Malik.
\newblock Learning category-specific mesh reconstruction from image
  collections.
\newblock In {\em ECCV}, 2018.

\bibitem{kato2018neural}
Hiroharu Kato, Yoshitaka Ushiku, and Tatsuya Harada.
\newblock Neural 3d mesh renderer.
\newblock In {\em Proceedings of the IEEE conference on computer vision and
  pattern recognition}, pages 3907--3916, 2018.

\bibitem{kim2020softflow}
Hyeongju Kim, Hyeonseung Lee, Woo~Hyun Kang, Joun~Yeop Lee, and Nam~Soo Kim.
\newblock Softflow: Probabilistic framework for normalizing flow on manifolds.
\newblock {\em Advances in Neural Information Processing Systems},
  33:16388--16397, 2020.

\bibitem{klokov2020discrete}
Roman Klokov, Edmond Boyer, and Jakob Verbeek.
\newblock Discrete point flow networks for efficient point cloud generation.
\newblock In {\em European Conference on Computer Vision}, pages 694--710.
  Springer, 2020.

\bibitem{kocabas2020vibe}
Muhammed Kocabas, Nikos Athanasiou, and Michael~J Black.
\newblock Vibe: Video inference for human body pose and shape estimation.
\newblock In {\em Proceedings of the IEEE/CVF conference on computer vision and
  pattern recognition}, pages 5253--5263, 2020.

\bibitem{kolotouros2019learning}
Nikos Kolotouros, Georgios Pavlakos, Michael~J Black, and Kostas Daniilidis.
\newblock Learning to reconstruct 3d human pose and shape via model-fitting in
  the loop.
\newblock In {\em Proceedings of the IEEE/CVF International Conference on
  Computer Vision}, pages 2252--2261, 2019.

\bibitem{kutuk2022semantic}
Z{\"u}lfiye K{\"u}t{\"u}k and G{\"o}rkem Algan.
\newblock Semantic segmentation for thermal images: A comparative survey.
\newblock In {\em Proceedings of the IEEE/CVF Conference on Computer Vision and
  Pattern Recognition}, pages 286--295, 2022.

\bibitem{li2020simple}
Jia Li, Wen Su, and Zengfu Wang.
\newblock Simple pose: Rethinking and improving a bottom-up approach for
  multi-person pose estimation.
\newblock In {\em Proceedings of the AAAI conference on artificial
  intelligence}, volume~34, pages 11354--11361, 2020.

\bibitem{li2021hybrik}
Jiefeng Li, Chao Xu, Zhicun Chen, Siyuan Bian, Lixin Yang, and Cewu Lu.
\newblock Hybrik: A hybrid analytical-neural inverse kinematics solution for 3d
  human pose and shape estimation.
\newblock In {\em Proceedings of the IEEE/CVF Conference on Computer Vision and
  Pattern Recognition}, pages 3383--3393, 2021.

\bibitem{li2018removal}
Ning Li, Yongqiang Zhao, Quan Pan, and Seong~G Kong.
\newblock Removal of reflections in lwir image with polarization
  characteristics.
\newblock {\em Optics express}, 26(13):16488--16504, 2018.

\bibitem{li2018differentiable}
Tzu-Mao Li, Miika Aittala, Fr{\'e}do Durand, and Jaakko Lehtinen.
\newblock Differentiable monte carlo ray tracing through edge sampling.
\newblock {\em ACM Transactions on Graphics (TOG)}, 37(6):1--11, 2018.

\bibitem{li2020differentiable}
Tzu-Mao Li, Michal Luk{\'a}{\v{c}}, Micha{\"e}l Gharbi, and Jonathan
  Ragan-Kelley.
\newblock Differentiable vector graphics rasterization for editing and
  learning.
\newblock {\em ACM Transactions on Graphics (TOG)}, 39(6):1--15, 2020.

\bibitem{li2020self}
Xueting Li, Sifei Liu, Kihwan Kim, Shalini~De Mello, Varun Jampani, Ming-Hsuan
  Yang, and Jan Kautz.
\newblock Self-supervised single-view 3d reconstruction via semantic
  consistency.
\newblock In {\em European Conference on Computer Vision}, pages 677--693.
  Springer, 2020.

\bibitem{lin2021end}
Kevin Lin, Lijuan Wang, and Zicheng Liu.
\newblock End-to-end human pose and mesh reconstruction with transformers.
\newblock In {\em Proceedings of the IEEE/CVF Conference on Computer Vision and
  Pattern Recognition}, pages 1954--1963, 2021.

\bibitem{liu2022shadows}
Ruoshi Liu, Sachit Menon, Chengzhi Mao, Dennis Park, Simon Stent, and Carl
  Vondrick.
\newblock Shadows shed light on 3d objects.
\newblock {\em arXiv preprint arXiv:2206.08990}, 2022.

\bibitem{liu2019soft}
Shichen Liu, Tianye Li, Weikai Chen, and Hao Li.
\newblock Soft rasterizer: A differentiable renderer for image-based 3d
  reasoning.
\newblock In {\em Proceedings of the IEEE/CVF International Conference on
  Computer Vision}, pages 7708--7717, 2019.

\bibitem{loper2014opendr}
Matthew~M Loper and Michael~J Black.
\newblock Opendr: An approximate differentiable renderer.
\newblock In {\em European Conference on Computer Vision}, pages 154--169.
  Springer, 2014.

\bibitem{luo2021diffusion}
Shitong Luo and Wei Hu.
\newblock Diffusion probabilistic models for 3d point cloud generation.
\newblock In {\em Proceedings of the IEEE/CVF Conference on Computer Vision and
  Pattern Recognition}, pages 2837--2845, 2021.

\bibitem{maeda2019thermal}
Tomohiro Maeda, Yiqin Wang, Ramesh Raskar, and Achuta Kadambi.
\newblock Thermal non-line-of-sight imaging.
\newblock In {\em 2019 IEEE International Conference on Computational
  Photography (ICCP)}, pages 1--11. IEEE, 2019.

\bibitem{maset2017photogrammetric}
E Maset, A Fusiello, F Crosilla, R Toldo, and D Zorzetto.
\newblock Photogrammetric 3d building reconstruction from thermal images.
\newblock {\em ISPRS Annals of the Photogrammetry, Remote Sensing and Spatial
  Information Sciences}, 4:25, 2017.

\bibitem{mescheder2019occupancy}
Lars Mescheder, Michael Oechsle, Michael Niemeyer, Sebastian Nowozin, and
  Andreas Geiger.
\newblock Occupancy networks: Learning 3d reconstruction in function space.
\newblock In {\em Proceedings of the IEEE/CVF Conference on Computer Vision and
  Pattern Recognition}, pages 4460--4470, 2019.

\bibitem{mildenhall2021nerf}
Ben Mildenhall, Pratul~P Srinivasan, Matthew Tancik, Jonathan~T Barron, Ravi
  Ramamoorthi, and Ren Ng.
\newblock Nerf: Representing scenes as neural radiance fields for view
  synthesis.
\newblock {\em Communications of the ACM}, 65(1):99--106, 2021.

\bibitem{moon2020i2l}
Gyeongsik Moon and Kyoung~Mu Lee.
\newblock I2l-meshnet: Image-to-lixel prediction network for accurate 3d human
  pose and mesh estimation from a single rgb image.
\newblock In {\em European Conference on Computer Vision}, pages 752--768.
  Springer, 2020.

\bibitem{muller2022instant}
Thomas M{\"u}ller, Alex Evans, Christoph Schied, and Alexander Keller.
\newblock Instant neural graphics primitives with a multiresolution hash
  encoding.
\newblock {\em arXiv preprint arXiv:2201.05989}, 2022.

\bibitem{nie2019single}
Xuecheng Nie, Jiashi Feng, Jianfeng Zhang, and Shuicheng Yan.
\newblock Single-stage multi-person pose machines.
\newblock In {\em Proceedings of the IEEE/CVF international conference on
  computer vision}, pages 6951--6960, 2019.

\bibitem{niemeyer2021giraffe}
Michael Niemeyer and Andreas Geiger.
\newblock Giraffe: Representing scenes as compositional generative neural
  feature fields.
\newblock In {\em Proceedings of the IEEE/CVF Conference on Computer Vision and
  Pattern Recognition}, pages 11453--11464, 2021.

\bibitem{nimier2019mitsuba}
Merlin Nimier-David, Delio Vicini, Tizian Zeltner, and Wenzel Jakob.
\newblock Mitsuba 2: A retargetable forward and inverse renderer.
\newblock {\em ACM Transactions on Graphics (TOG)}, 38(6):1--17, 2019.

\bibitem{oren1994generalization}
Michael Oren and Shree~K Nayar.
\newblock Generalization of lambert's reflectance model.
\newblock In {\em Proceedings of the 21st annual conference on Computer
  graphics and interactive techniques}, pages 239--246, 1994.

\bibitem{park2019deepsdf}
Jeong~Joon Park, Peter Florence, Julian Straub, Richard Newcombe, and Steven
  Lovegrove.
\newblock Deepsdf: Learning continuous signed distance functions for shape
  representation.
\newblock In {\em Proceedings of the IEEE/CVF conference on computer vision and
  pattern recognition}, pages 165--174, 2019.

\bibitem{park2021nerfies}
Keunhong Park, Utkarsh Sinha, Jonathan~T Barron, Sofien Bouaziz, Dan~B Goldman,
  Steven~M Seitz, and Ricardo Martin-Brualla.
\newblock Nerfies: Deformable neural radiance fields.
\newblock In {\em Proceedings of the IEEE/CVF International Conference on
  Computer Vision}, pages 5865--5874, 2021.

\bibitem{pavlakos2019expressive}
Georgios Pavlakos, Vasileios Choutas, Nima Ghorbani, Timo Bolkart, Ahmed~AA
  Osman, Dimitrios Tzionas, and Michael~J Black.
\newblock Expressive body capture: 3d hands, face, and body from a single
  image.
\newblock In {\em Proceedings of the IEEE/CVF conference on computer vision and
  pattern recognition}, pages 10975--10985, 2019.

\bibitem{ramasinghe2020spectral}
Sameera Ramasinghe, Salman Khan, Nick Barnes, and Stephen Gould.
\newblock Spectral-gans for high-resolution 3d point-cloud generation.
\newblock In {\em 2020 IEEE/RSJ International Conference on Intelligent Robots
  and Systems (IROS)}, pages 8169--8176. IEEE, 2020.

\bibitem{ravi2020accelerating}
Nikhila Ravi, Jeremy Reizenstein, David Novotny, Taylor Gordon, Wan-Yen Lo,
  Justin Johnson, and Georgia Gkioxari.
\newblock Accelerating 3d deep learning with pytorch3d.
\newblock {\em arXiv preprint arXiv:2007.08501}, 2020.

\bibitem{rempe2021humor}
Davis Rempe, Tolga Birdal, Aaron Hertzmann, Jimei Yang, Srinath Sridhar, and
  Leonidas~J Guibas.
\newblock Humor: 3d human motion model for robust pose estimation.
\newblock In {\em Proceedings of the IEEE/CVF International Conference on
  Computer Vision}, pages 11488--11499, 2021.

\bibitem{rivadeneira2022thermal}
Rafael~E Rivadeneira, Angel~D Sappa, Boris~X Vintimilla, Jin Kim, Dogun Kim,
  Zhihao Li, Yingchun Jian, Bo Yan, Leilei Cao, Fengliang Qi, et~al.
\newblock Thermal image super-resolution challenge results-pbvs 2022.
\newblock In {\em Proceedings of the IEEE/CVF Conference on Computer Vision and
  Pattern Recognition}, pages 418--426, 2022.

\bibitem{sage20203d}
Agata Sage, Daniel Ledwo{\'n}, Jan Juszczyk, and Pawe{\l} Badura.
\newblock 3d thermal volume reconstruction from 2d infrared images—a
  preliminary study.
\newblock In {\em International Scientific Conference Advances in Applied
  Biomechanics}, pages 371--379. Springer, 2020.

\bibitem{schramm2022combining}
Sebastian Schramm, Phil Osterhold, Robert Schmoll, and Andreas Kroll.
\newblock Combining modern 3d reconstruction and thermal imaging: Generation of
  large-scale 3d thermograms in real-time.
\newblock {\em Quantitative InfraRed Thermography Journal}, 19(5):295--311,
  2022.

\bibitem{sigal2010humaneva}
Leonid Sigal, Alexandru~O Balan, and Michael~J Black.
\newblock Humaneva: Synchronized video and motion capture dataset and baseline
  algorithm for evaluation of articulated human motion.
\newblock {\em International journal of computer vision}, 87(1):4--27, 2010.

\bibitem{srinivasan2021nerv}
Pratul~P Srinivasan, Boyang Deng, Xiuming Zhang, Matthew Tancik, Ben
  @article{li2018differentiable, title={Differentiable monte carlo ray tracing
  through edge sampling}, author={Li, Tzu-Mao and Aittala, Miika and Durand,
  Fr{\'e}do and Lehtinen, Jaakko}, journal={ACM Transactions on Graphics
  (TOG)}, volume={37}, number={6}, pages={1--11}, year={2018}, publisher={ACM
  New York, NY, USA} }Mildenhall, and Jonathan~T Barron.
\newblock Nerv: Neural reflectance and visibility fields for relighting and
  view synthesis.
\newblock In {\em Proceedings of the IEEE/CVF Conference on Computer Vision and
  Pattern Recognition}, pages 7495--7504, 2021.

\bibitem{sun2019deep}
Ke Sun, Bin Xiao, Dong Liu, and Jingdong Wang.
\newblock Deep high-resolution representation learning for human pose
  estimation.
\newblock In {\em Proceedings of the IEEE/CVF conference on computer vision and
  pattern recognition}, pages 5693--5703, 2019.

\bibitem{tiwari2022pose}
Garvita Tiwari, Dimitrije Anti{\'c}, Jan~Eric Lenssen, Nikolaos Sarafianos,
  Tony Tung, and Gerard Pons-Moll.
\newblock Pose-ndf: Modeling human pose manifolds with neural distance fields.
\newblock In {\em European Conference on Computer Vision}, pages 572--589.
  Springer, 2022.

\bibitem{treptow2006real}
Andre Treptow, Grzegorz Cielniak, and Tom Duckett.
\newblock Real-time people tracking for mobile robots using thermal vision.
\newblock {\em Robotics and Autonomous Systems}, 54(9):729--739, 2006.

\bibitem{vicini2022differentiable}
Delio Vicini, S{\'e}bastien Speierer, and Wenzel Jakob.
\newblock Differentiable signed distance function rendering.
\newblock {\em ACM Transactions on Graphics (TOG)}, 41(4):1--18, 2022.

\bibitem{wu2016learning}
Jiajun Wu, Chengkai Zhang, Tianfan Xue, Bill Freeman, and Josh Tenenbaum.
\newblock Learning a probabilistic latent space of object shapes via 3d
  generative-adversarial modeling.
\newblock {\em Advances in neural information processing systems}, 29, 2016.

\bibitem{wu2020unsupervised}
Shangzhe Wu, Christian Rupprecht, and Andrea Vedaldi.
\newblock Unsupervised learning of probably symmetric deformable 3d objects
  from images in the wild.
\newblock In {\em Proceedings of the IEEE/CVF Conference on Computer Vision and
  Pattern Recognition}, pages 1--10, 2020.

\bibitem{wu2019sagnet}
Zhijie Wu, Xiang Wang, Di Lin, Dani Lischinski, Daniel Cohen-Or, and Hui Huang.
\newblock Sagnet: Structure-aware generative network for 3d-shape modeling.
\newblock {\em ACM Transactions on Graphics (TOG)}, 38(4):1--14, 2019.

\bibitem{ye2021shelf}
Yufei Ye, Shubham Tulsiani, and Abhinav Gupta.
\newblock Shelf-supervised mesh prediction in the wild.
\newblock In {\em Proceedings of the IEEE/CVF Conference on Computer Vision and
  Pattern Recognition}, pages 8843--8852, 2021.

\bibitem{yu2016deep}
Xiang Yu, Feng Zhou, and Manmohan Chandraker.
\newblock Deep deformation network for object landmark localization.
\newblock In {\em European conference on computer vision}, pages 52--70.
  Springer, 2016.

\bibitem{zakharov2020autolabeling}
Sergey Zakharov, Wadim Kehl, Arjun Bhargava, and Adrien Gaidon.
\newblock Autolabeling 3d objects with differentiable rendering of sdf shape
  priors.
\newblock In {\em Proceedings of the IEEE/CVF Conference on Computer Vision and
  Pattern Recognition}, pages 12224--12233, 2020.

\bibitem{zeise2016temperature}
Bj{\"o}rn Zeise and Bernardo Wagner.
\newblock Temperature correction and reflection removal in thermal images using
  3d temperature mapping.
\newblock In {\em ICINCO (2)}, pages 158--165, 2016.

\bibitem{zhang2019differential}
Cheng Zhang, Lifan Wu, Changxi Zheng, Ioannis Gkioulekas, Ravi Ramamoorthi, and
  Shuang Zhao.
\newblock A differential theory of radiative transfer.
\newblock {\em ACM Transactions on Graphics (TOG)}, 38(6):1--16, 2019.

\end{thebibliography}
}

\end{document}